\documentclass{article}
\usepackage{spconf,amsmath,graphicx}
\usepackage{multirow}
\usepackage{array}
\usepackage{tikz}
\usetikzlibrary{shapes.geometric, arrows}
\tikzstyle{block1} = [rectangle, rounded corners, minimum width=1cm, minimum height=2cm,text centered, draw=black, fill=white!30, text width=2.15cm]
\tikzstyle{block2} = [rectangle, rounded corners, minimum width=1cm, minimum height=2cm,text centered, draw=black, fill=white!30, text width=1.8cm]
\tikzstyle{arrow} = [thick,->,>=stealth]

\title{Spatio-temporal Crop Classification on Volumetric data}
\name{Muhammad Usman Qadeer, Salar Saeed, Murtaza Taj and Abubakr Muhammad}
\address{}
\newcommand\Tstrut{\rule{0pt}{2.6ex}}         
\newcommand\Bstrut{\rule[-0.9ex]{0pt}{0pt}}   
\begin{document}
	%
	\maketitle
	\vspace{-0.5cm}
	\begin{abstract}
		Large-area crop classification using multi-spectral imagery is a widely studied problem for several decades and is generally addressed using classical Random Forest classifier. Recently, deep convolutional neural networks (DCNN) have been proposed. However, these methods only achieved results comparable with Random Forest. In this work, we present a novel CNN based architecture for large-area crop classification. Our methodology combines both spatio-temporal analysis via 3D CNN as well as temporal analysis via 1D CNN. We evaluated the efficacy of our approach on Yolo and Imperial county benchmark datasets. Our combined strategy outperforms both classical as well as recent DCNN based methods in terms of classification accuracy by $2\%$ while maintaining a minimum number of parameters and the lowest inference time. 
		
	\end{abstract}
	\section{Introduction}
	\label{sec:intro}
	\vspace{-0.25cm}
	The freely available multispectral satellite imagery and advancement in modern machine learning have paved the way for a wide variety of applications. These include disaster assessment~\cite{dao2015object}, crop classification~\cite{csillik2019object, kobayashi2020crop}, urbanization~\cite{deepCountISPRS19} and environment monitoring~\cite{donlon2012global}. 
	The increasing quality and resolution of available remote sensing imagery have made it possible to perform robust crop monitoring and yield estimation over large areas~\cite{battude2016estimating}. For instance, the Sentinel-2 satellite imagery with a spatial resolution of 10m and Landsat-8 imagery with a spatial resolution of 30m is available every five days and 16 days, respectively. Thus, enabling us to perform detailed spatio-temporal analysis~\cite{vuolo2018much}. 
	\par
	Supervised learning is considered to be the state-of-the-art approach to produce crop maps. Some traditional machine learning algorithms like Support Vector Machines (SVM) \cite{saini2018crop}, K-nearest neighbors (KNN) \cite{thanh2018comparison}, Cart \cite{sonobe2017experimental}, and Random Forest \cite{rodriguez2012assessment} have been widely applied for crop classification. Among these methods, those using a series of images at different time stamps~\cite{gomez2016optical} have shown better results than the one using a single image \cite{saini2018crop}.
	
	\par
	Due to the popularity of deep learning, recently, the problem of crop classification has also been attempted using different deep learning architectures. In earlier work, a 1D convolutional neural network model has been proposed~\cite{zhou2018crops} which stacked features of different time stamps as in the case of random forest.  To use the temporal information more efficiently, ~\cite{pelletier2019temporal} has proposed a new 1D CNN model. This method feeds multivariate time series to 1D CNN instead of stacked values of features. Similarly, a new 1D CNN architecture based on the idea of the Inception network has been introduced~\cite{zhong2019deep}.  It has shown improved performance on traditional machine learning algorithms, including Random Forest and XGboost~\cite{abdi2020land}.
	\par
	The 2D convolutional neural networks show outstanding image classification performance, and many architectures, including Alexnet~\cite{krizhevsky2012imagenet}, Inception, Resnets~\cite{he2016deep}, etc., have been introduced. A new dataset comprising Sentinel-2 images of different landcover types from around the world has been developed~\cite{helber2019eurosat} and different 2D CNN architectures, including Resnet50 and GoogleNet~\cite{szegedy2015going}, have been compared to classify the multispectral images. Recently, an ensemble of 2D CNN has been used for crop classification using multi-temporal Sentinel-1 and Landsat-8 imagery~\cite{kussul2017deep}. This study shows the significance of using spatial information in addition to temporal one.
	\par
	The recurrent neural networks are very popular for applications where time-series data is used. The idea of using LSTM \cite{zhou2019long} has also been applied to landcover classification using multi-temporal SAR imagery~\cite{zhong2019deep}. This work has claimed that LSTM performs worse than Random Forest classifier for crop classification using multispectral imagery. The recurrent convolutional neural network (R-CNN)~\cite{rcnn} uses LSTM followed by 2D CNN, but unlike~\cite{kussul2017deep}, this architecture uses the information of only one pixel. The idea of combining CNN and LSTM has also been introduced for landcover classification~\cite{kwak2019combining}\cite{russwurm2018multi}.
	\par
	The 3D convolutional neural networks are widely considered for applications related to video, medical imaging, and remote sensing~\cite{BhimraICASSP2019}. In \cite{ji20183d}, a 3D CNN architecture has been presented showing better performance than 2D CNN as it uses the temporal information better than the 2D CNN where timestamps of all features are stacked on one axis.
	
	\begin{figure*}[t]
		\centering
		\includegraphics[width = 18cm]{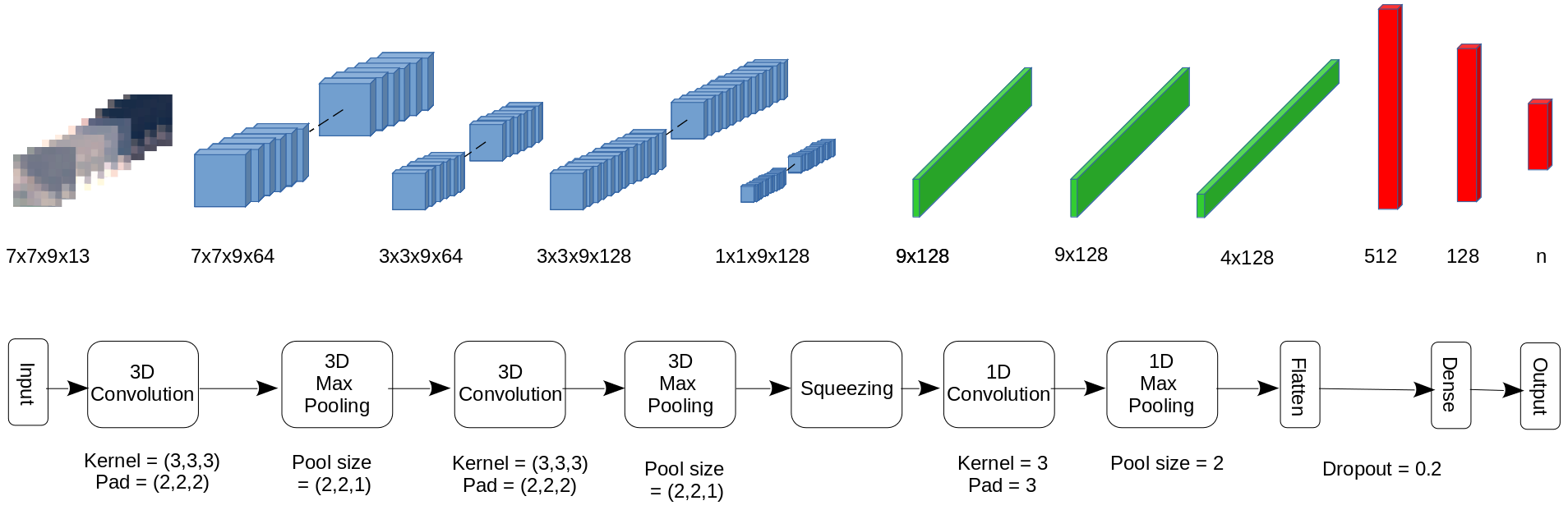}
		\caption{Proposed model architectures showing combination of spatio-temporal and temporal convolution blocks. Spatio-temporal block via 3D CNN are shown in blue whereas the temporal only block via 1D convolution are shown in green.}
		\vspace{-0.4cm}
		\label{fig:model}
	\end{figure*}
	
	Existing literature suggests various ways to exploit available spatio-temporal data~\cite{BhimraICASSP2019,ji20183d} however, these methods are either good at utilizing spatial information or temporal information but not both. We instead propose a novel DCNN based architecture that combines both spatial as well as temporal analysis. In the first stage of our architecture, we propose to use 3D convolutions that perform spatio-temporal analysis without collapsing the temporal dimension. Once the spatio-temporal features are extracted via multiple 3D CNN layers, we introduce temporal only analysis to further extract important information from the temporal dimension only. This spatio-temporal followed by temporal only analysis helps eliminate noise, usually present in the temporal only analysis. Thus the proposed approach outperforms classical as well as state-of-art-methods on benchmark datasets.

	\section{Methodology}
	\label{sec:method}
	\vspace{-0.25cm}
	
	\subsection{Data Pre-processing}
	\vspace{-0.2cm}
	We have divided the pre-processing of data into five steps. The first step is to select all the images of a multispectral satellite for the region of interest in a suitable period, covering every crop's cropping season under consideration. For this purpose, the NDVI time series for different crops have been examined. The second step is to select the least cloudy images in that period. In our case, the images with cloud cover less than $10\%$ have been selected. The third step is the cloud masking of the chosen images. The fourth step is to take the median of images chosen for each month. This step is for larger areas, as in our case, because the region can't be covered in a single image of a satellite. In the fifth and final step, the missing pixels due to the cloud or other reasons have been filled, which can be done using simple interpolation or linear regression. All these pre-processing steps have been performed on the Google Earth Engine Platform.
	
	%
	\vspace{-0.2cm}
	\subsection{Feature Selection}
	\vspace{-0.2cm}
	For each image in our dataset, $13$ features have been selected. The selected features have six multi-spectral satellite imagery bands, including Blue, Green, Red, Near Infrared (NIR), and two Short Wave Infrared (SWIR) bands. Table~\ref{tab:bands} shows the selected Sentinel-2 and Landsat-8 bands. The other seven features are indices derived from these bands, including Normalized Difference Vegetation Index (NDVI), Enhanced Vegetation Index, Green Normalized Difference Vegetation Index (GNDVI), Soil Adjusted Vegetation Index (SAVI), Bare Soil Index (BSI), Normalized Difference Water Index (NDWI) and Normalized Difference Buildup Index (NDBI). These indices~\cite{pal2017comparison,vermote2016preliminary} are calculated as follows:
	\begin{table}[b]
		\centering
		\caption{Selected Bands of Landsat-8 and Sentinel-2\Bstrut}
		\begin{tabular}{>{\centering}p{1.6cm} >{\centering} p{0.5cm}>{\centering}p{0.8cm}>{\centering}p{0.6cm}>{\centering}p{0.6cm}p{0.8cm}p{0.8cm}} 
			\hline 
			\textbf{Bands}&\textbf{Blue}&\textbf{Green}&\textbf{Red}&\textbf{NIR}&\textbf{SWIR1}& \textbf{SWIR2}\Tstrut\Bstrut\\
			\hline\hline
			\textbf{Sentinel-2}&B2&B3&B4&B8&B11&B12\Tstrut\\
			\textbf{Landsat-8}&B2&B3&B4&B5&B6&B7\\
			\hline
		\end{tabular}
		\label{tab:bands}
	\end{table}

	\begin{center}
		$\text{NDVI} = \frac{NIR - RED}{NIR + RED}~~~~,~~\text{GNDVI} = \frac{NIR - GREEN}{NIR + GREEN}$
	\end{center}
	\begin{center}
		$\text{EVI} = 2.5\times\frac{NIR - RED}{NIR + 6\times RED-7.5\times BLUE + 1}$
	\end{center}
	\begin{center}
		$\text{SAVI} = 1.5\times\frac{NIR - RED}{NIR + RED + 0.5}$
	\end{center}
	\begin{center}
		$\text{BSI} = \frac{(SWIR + RED) - (NIR + BLUE)}{(SWIR + RED) + (NIR + BLUE)}$
	\end{center}
	\begin{center}
		$\text{NDBI} = \frac{SWIR1 - NIR}{SWIR1 + NIR}~~,~~~\text{NDWI}= \frac{GREEN - NIR}{GREEN + NIR}$
	\end{center}
	\subsection{Model Architecture}
	\vspace{-0.2cm}
	The model architecture is shown in Fig.~\ref{fig:model}. The model consists of three parts concatenated to each other. The first part shown in blue performs spatio-temporal convolution using 3D CNN. The second part shown in green performs the temporal convolution. The output of 3D CNN is squeezed before feeding to 1D CNN. These two parts extract the features from input. The third part consists of a fully connected neural network, which predicts the label from multi-temporal input images.
	\par
	The model input shape is $(r, r, t, c)$ where $r$ is the spatial dimension of input, $t$ is the number of timestamps, and $c$ is the number of channels. In our implementation, spatial dimension $r$ is $7$, timestamps $t$ are $9$ and channels $c$ are $13$. Each convolutional block has a batch normalization layer before activation.  It has been observed that with batch-normalization, the model converges faster as well as shows better accuracy.

	\section{RESULTS AND EVALUATION}
	\label{sec:results}
	\vspace{-0.2cm}
	\subsection{Experimental Setup}
	\vspace{-0.2cm}
	The experiments have been performed on two different California counties, including Imperial County and Yolo County. For both regions, the crop survey data provided by the California Department of Water Resources (CDWR) \cite{dataset} for the year 2016 have been used as ground truth information.
	\begin{figure}[t]
		\centering
		\includegraphics[width=8.5cm]{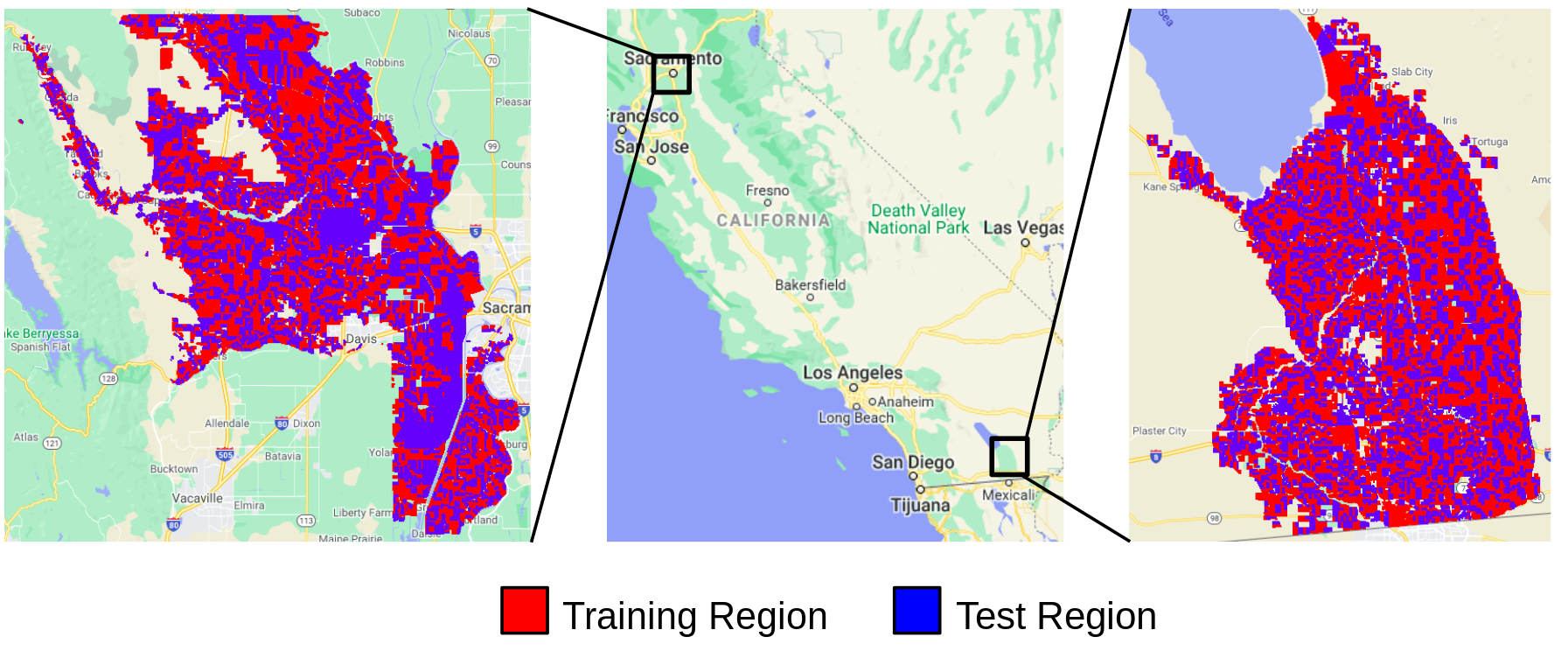}
		\caption{The regions of interest showing training (red) and test (blue) data distribution for Yolo (left) and Imperial Counties (right) distribution.}
		\vspace{-0.2cm}
		\label{fig:mapsamples}
	\end{figure}
	\par
	For Imperial County, $10$ major classes have been considered, while $14$ classes have been taken into account for Yolo County. The "Others" class contains either minor crop types, fallow land, or build up. There are about $7256$ and $7340$ total parcels for Imperial County and Yolo County, respectively, as mentioned in Table~\ref{tab:yolo_samples}. These parcels have been split into $60\%$ training data and $40\%$ test data. The fields have been divided into training and test sets before sampling to ensure that no sample from the training region is included in the test set. The Fig.~\ref{fig:mapsamples} shows the distribution of training and testing parcels. The imagery of Sentinel-2 and Landsat-8 has been used for Imperial County and Yolo County, respectively. In both cases, nine scenes from March to November 2016 have been used. The same dataset has been used to train all the classifiers, including proposed and state-of-the-art methods, in both pixel and parcel-based analysis. 
	
	\begin{table}[ht]
		\centering
		\small
		\caption{Summary of survey data from Imperial and Yolo counties provided by California Department of Water Resources (CDWR)}
		\begin{tabular}{p{1.6cm} p{0.8cm} p{0.8cm} | p{1.6cm} p{0.8cm} p{0.8cm}}
			
			\hline 
			\multicolumn{3}{c}{\textbf{Imperial County}} & \multicolumn{3}{c}{\textbf{Yolo County}}\Tstrut\Bstrut\\
			\hline
			\multirow{2}{*}{\textbf{Class}} &\textbf{Parcels} & \textbf{\%age} & \multirow{2}{*}{\textbf{Class}} &\textbf{Parcels} & \textbf{\%age}\Tstrut\\ 
			&  \textbf{Count} & \textbf{area} & &  \textbf{Count} & \textbf{area}\Bstrut\\
			
			\hline \hline
			Alfalfa & 2156 & 31.3\% & Rice & 350 & 11.2\%\Tstrut\\
			Pastures & 1150 & 15.4\%& Safflower & 161 & 2.4\%\\
			Lettuce& 619 & 8.2\%& Corn & 150 & 2.1\%\\
			Wheat & 315 & 4.3\%& Field crops & 407 & 6.6\%\\
			Onions & 229& 3.0\% & Alfalfa & 575 & 8.7\%\\
			Truck Crops & 1026 & 11.6\% & Pastures & 440 & 4.3\%\\
			Corn & 232& 3.5\% & Cucurbits & 168 & 1.8\%\\
			Field Crops & 192 & 2.9\% & Tomatoes & 496 & 9.8\%\\
			Subtropical &271 & 1.6\% & Truck Crops & 289 & 1.3\%\\
			Others & 1066 & 18.2\%& Almonds & 799 & 8.9\%\\
			& & & Deciduous & 867 & 5.3\%\\
			& & & Subtropical & 176 & 1.1\%\\
			& & & Vineyard & 688 & 5.6\%\\
			& & & Others & 1874 & 30.9\%\\\hline
		\end{tabular}
		\vspace{-0.4cm}
		\label{tab:yolo_samples}
	\end{table}
	
	\subsection{Comparative Analysis}
	\vspace{-0.2cm}
	We compared our proposed hybrid model with four state-of-the-art methods, namely Random Forest, 1D CNN \cite{zhong2019deep}, 2D CNN\cite{kussul2017deep} and 3D CNN\cite{ji20183d}
	
	\begin{table*}[t]
		\centering
		\small
		\caption{Table showing comparison of our proposed 3D$\rightarrow$1D CNN with state-of-the-art methods on Imperial and Yolo counties. The comparison is performed using pixel-wise and parcel-wise strategies (provides single classification within parcel boundary)}
		\resizebox{0.9\textwidth}{!}{
			\begin{tabular}
				{c|cc|cc|cc|cc}
				\hline 
				\multirow{3}{*}{\textbf{Classifier}}&\multicolumn{4}{c}{\textbf{Imperial County}} & \multicolumn{4}{c}{\textbf{Yolo County}}\Tstrut\Bstrut\\
				\cline{2-9}
				&\multicolumn{2}{c|}{\textbf{Pixel-wise}} & \multicolumn{2}{c|}{\textbf{Parcel-wise}}&\multicolumn{2}{c|}{\textbf{Pixel-wise}} & \multicolumn{2}{c}{\textbf{Parcel-wise}}\Tstrut\Bstrut\\
				\cline{2-9}
				& \textbf{Accuracy}&\textbf{F1-Score}&\textbf{Accuracy}&\textbf{F1-Score}& \textbf{Accuracy}&\textbf{F1-Score}&\textbf{Accuracy}&\textbf{F1-Score}\Tstrut\Bstrut\\
				\hline \hline
				Random Forest & 78.96\% & 78.60\% & 80.61\% & 80.39\% & 86.68\% & 86.25\%& 82.23\% & 81.21\%\Tstrut\\
				1D CNN \cite{zhong2019deep} & 79.46\% & 79.05\%& 80.33\% & 80.24\%& 87.94\% & 87.80\%& 84.45\% & 83.90\%\\
				2D CNN \cite{kussul2017deep} & 79.71\% & 79.48\%& 80.54\% & 80.40\% & 88.50\% & 88.36\%& 84.55\% & 84.18\%\\
				3D CNN \cite{ji20183d} & 79.95\% & 79.70\%& 80.80\% & 80.72\%&88.01\% & 87.82\%& 83.17\% & 82.74\%\\
				3D$\rightarrow$1D CNN  (Ours)& \textbf{81.23\%} & \textbf{81.16\%}& \textbf{81.76\%} & \textbf{81.76\%} & \textbf{90.23\%} & \textbf{90.08\%} & \textbf{85.87\%} & \textbf{85.51\%}\\	
				\hline
				
			\end{tabular}
		}
		\label{tab:results}
	\end{table*}

	\begin{figure*}[ht]
		\centering
		\begin{minipage}[b]{0.3\linewidth}
			\centering
			\centerline{\includegraphics[width=5cm]{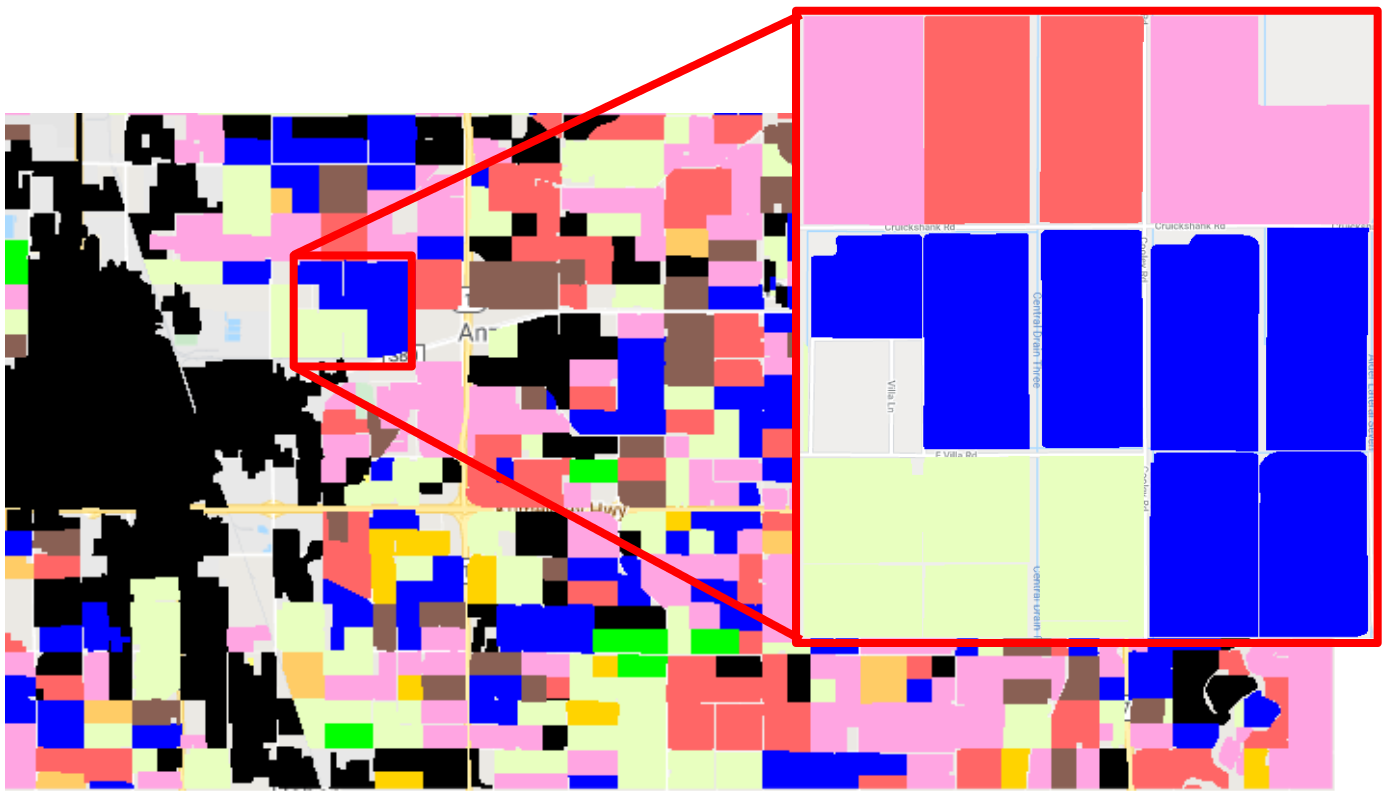}}
			\centerline{}\medskip
		\end{minipage}
		\begin{minipage}[b]{0.3\linewidth}
			\centering
			\centerline{\includegraphics[width=5cm]{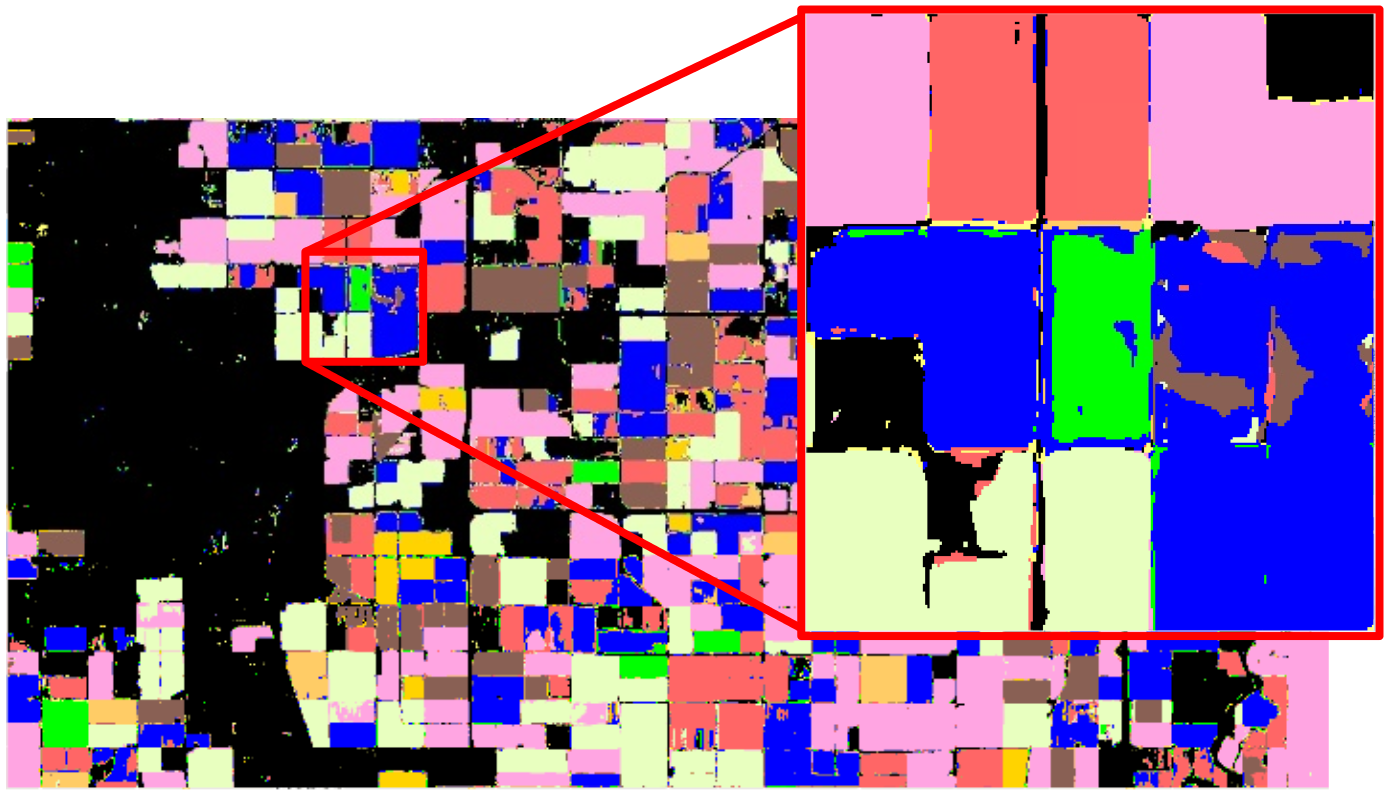}}
			\centerline{}\medskip
		\end{minipage}
		\begin{minipage}[b]{0.3\linewidth}
			\centering
			\centerline{\includegraphics[width=5cm]{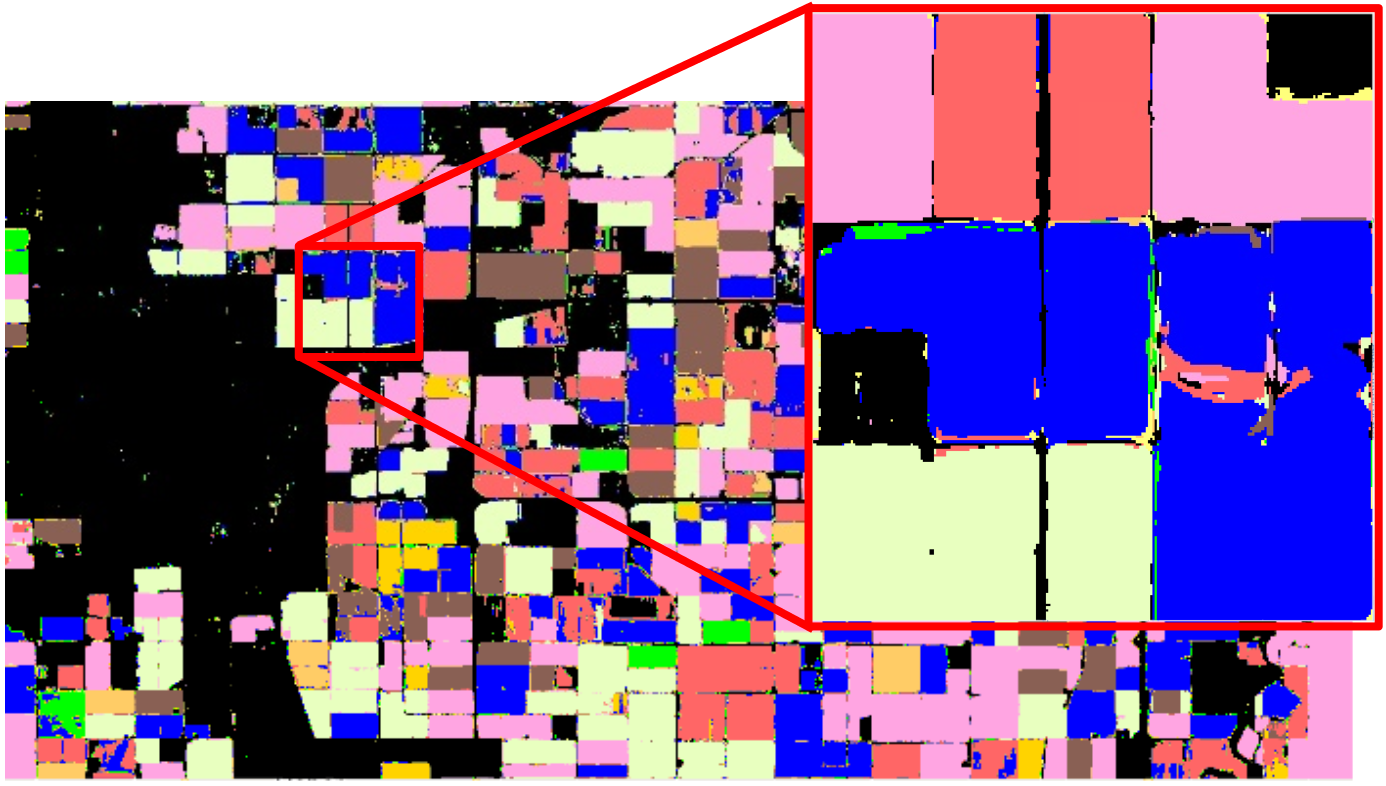}}
			\centerline{}\medskip
		\end{minipage}
		\begin{minipage}[b]{0.3\linewidth}
			\centering
			\centerline{\scalebox{-1}[1]{\includegraphics[width=5cm]{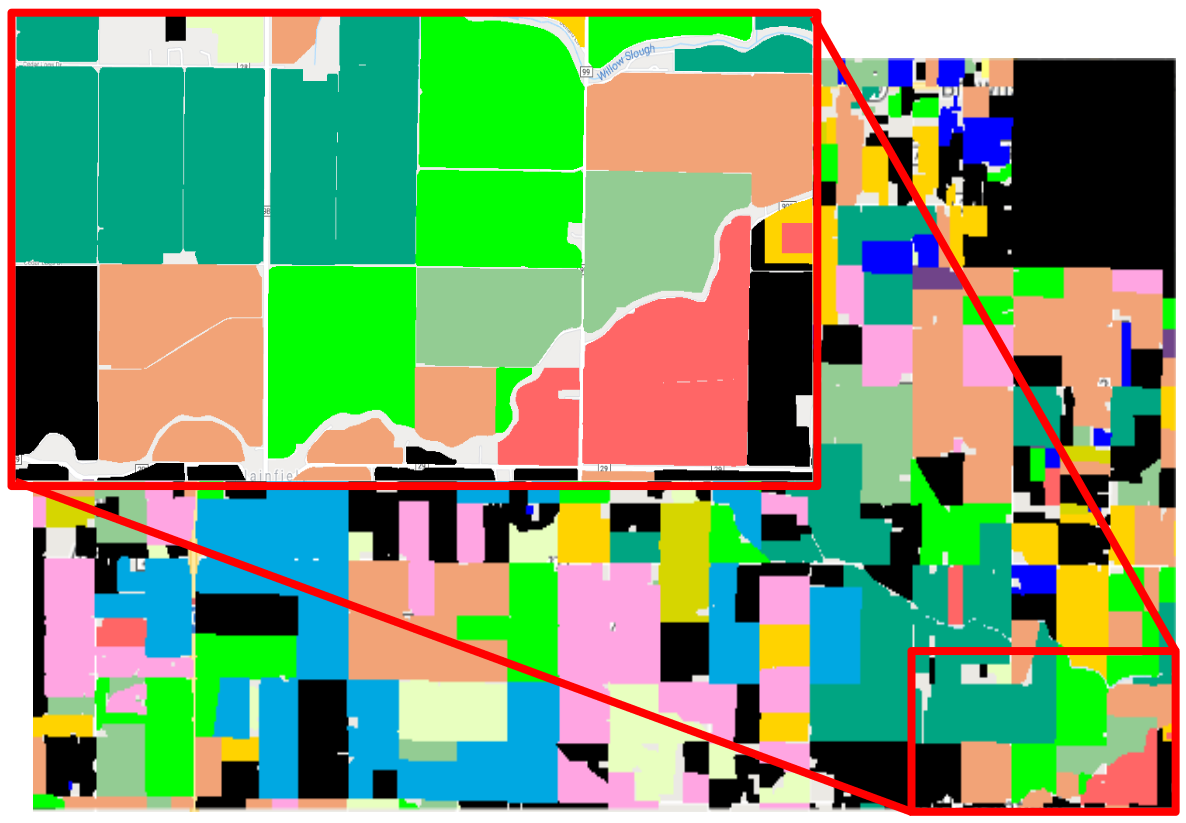}}}
			\centerline{(a) Ground Truth}\medskip
		\end{minipage}
		\begin{minipage}[b]{0.3\linewidth}
			\centering
			\centerline{\scalebox{-1}[1]{\includegraphics[width=5cm]{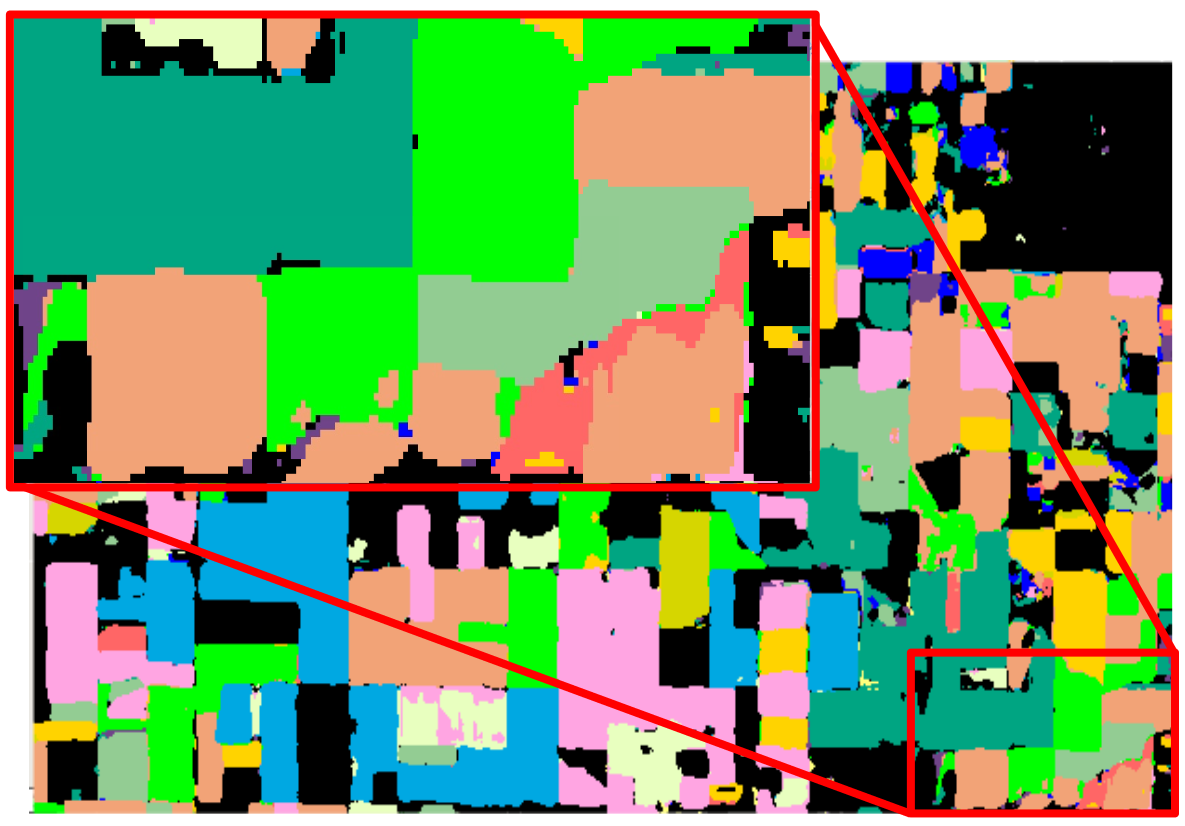}}}
			\centerline{(b) 3D CNN\cite{ji20183d}}\medskip
		\end{minipage}
		\begin{minipage}[b]{0.3\linewidth}
			\centering
			\centerline{\scalebox{-1}[1]{\includegraphics[width=5cm]{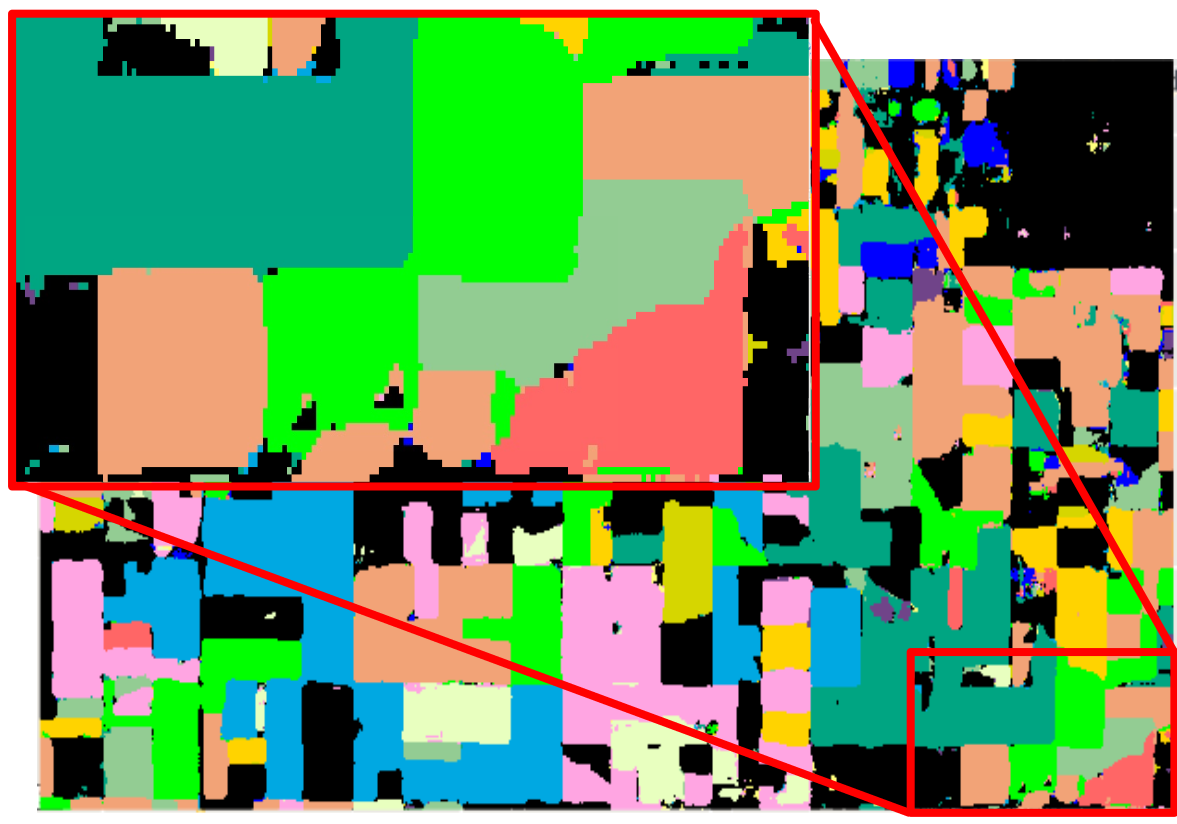}}}
			\centerline{(c) 3D$\rightarrow$1D CNN (Ours)}\medskip
		\end{minipage}\vspace{-0.2cm}
		
		\caption{Ground truth and maps produced for Imperial County (Top) and Yolo County (Bottom) by 3D CNN and our network.\label{fig:maps}}
		\vspace{-0.2cm}

	\end{figure*}

	For random forest, the number of trees has been set to $100$, the minimum number of samples required to split a node is two, and the maximum number of features is set to the square root of the number of features. For the 1D Convolutional Neural Network, the architecture proposed in \cite{zhong2019deep} has been used except that more features have been used instead of just EVI. The idea given by \cite{pelletier2019temporal} has been applied to consider more features in 1D CNN. For 2D Convolutional Neural Network, five ensembles of 2D CNN have been used as proposed in \cite{kussul2017deep}. The number of filters in the network ensembles is 160, 170, 180, 190, and 200. Due to the larger number of input features, more filters have to be used. For 3D Convolutional Neural Network, the model proposed in \cite{ji20183d} using the concept of VGG has been used, but the number of filters has been increased to 64 and 128 in both layers. Our proposed method has the same architecture as explained in Sec.~\ref{sec:method}.
	\par
	All the deep learning networks have been trained for about $50$ epochs using Adam optimizer with a default learning rate of $0.001$. The batch size of $128$ has been used, and each network's best results have been recorded. The accuracy and weighted F1-score for each classifier are mentioned in the Tables~\ref{tab:results}. The table shows the pixel-wise accuracy and F1-score as well as the parcel-wise results. It is evident from both tables that our proposed model has outperformed all the other classifiers in both experiments. Also, other deep learning models don't show much better accuracy than the random forest, and parcel-wise accuracy for the Imperial county of random forest is greater than other deep learning models except for our proposed one. On the other hand, our 3D$\rightarrow$1D CNN model improves the accuracy and F1-score by $2\%$. The parcel-wise results have been calculated by assigning each parcel a label based on the majority voting of pixels in it. 
	\par
	Contrary to our expectation, the parcel-wise accuracy is $4-5\%$ less than pixel-wise accuracy for each classifier in the case of Yolo County. The Yolo County survey data analysis revealed that $3.5\%$ parcels in test data were tagged as multi-use, i.e., more than one crop was sown in those fields. In contrast, Imperial County data have no field labeled as multi-use. Therefore, accuracy increased in the case of Imperial County while decreased in the case of Yolo County. Another reason is the presence of larger polygons in the test set of Yolo County.
	\par
	
	The \textbf{Qualitative Analysis} is shown in Fig.~\ref{fig:maps} which shows the original labels from CDWR data and results produced by 3D CNN \cite{ji20183d} and our proposed 3D$\rightarrow$1D CNN model. In the figure, there are some blank spaces where ground truth was not available, and it can be observed that those blank spaces have been predicted as "Others" represented by black color. The zoomed portion of the results in Fig.~\ref{fig:maps} further illustrates that the maps produced by our hybrid model have lesser noise compared to the ones produced by vanilla 3D CNN. 
	
	\begin{table}[tbh]
		\centering
		\caption{Performance Comparison of State of the art models}
		\resizebox{0.9\columnwidth}{!}{
			\begin{tabular}{c c c} 
				\hline 
				\textbf{Classifier} & \textbf{Parameters} & \textbf{Inference time (ms)}\Tstrut\Bstrut\\
				\hline \hline
				1D CNN \cite{zhong2019deep} & 1,083,970  & 6.39\Tstrut\\ 
				2D CNN \cite{kussul2017deep}  & 4,278,282 & 7.11\\
				3D CNN \cite{ji20183d}  & 394,050  & 4.72\\
				3D$\rightarrow$1D CNN (Ours) & 361,406 &  4.58\\
				\hline
			\end{tabular}
		}
		\label{tab:performance}
	\end{table}
	
	\vspace{-0.4cm}
	\subsection{Computational Cost}
	\par
	Table~\ref{tab:performance} shows the comparison of the number of parameters in different deep learning models and their inference time. Our model has a lesser number of parameters as well as the fastest inference time. This is because, instead of flattening the output of 3D CNN, we fed it to 1D CNN, which reduced the number of parameters and improved the performance.

	\vspace{-0.2cm}
	\section{Conclusion}
	\vspace{-0.4cm}
	In this paper, a new architecture and a method for large-area crop classification have been presented. Our hybrid model combines the spatio-temporal representation via 3D CNN with temporal only representation via 1D CNN. Different neural network architectures proposed in the past for crop mapping have been implemented and compared with our designed model.  Experiments have been performed using multispectral imagery of different satellites to show the generality of our method. The experimentation has revealed that the proposed hybrid network performs better on benchmark CDWR datasets. 
	
	In the future, we aim to extend our approach to fields that have been marked as multi-use (having more than one crop in a small parcel) as multi-use is a more commonly observed pattern in agriculture-based economies, particularly among developing countries.

	\small
	\bibliographystyle{IEEEbib}
	
	\bibliography{refs}
	
\end{document}